\documentclass[twocolumn]{article}
\usepackage{graphicx}
\usepackage{amssymb}
\usepackage{amsmath}
\usepackage{epstopdf}
\usepackage{url}
\usepackage[margin=1.5cm]{geometry}
\usepackage[official]{eurosym}
\usepackage[font=small,labelfont=bf]{caption}

\newcommand{\J}{\mathbf{J}} 
\renewcommand{\a}{\mathbf{a}} 
\renewcommand{\o}{\mathbf{o}}

\newcommand{\W}{\mathbf{W}} 
 
\newcommand{\e}{\mathbf{e}} 
\newcommand{\s}{\mathbf{s}}

\newcommand{\f}{\mathbf{f}}

\newcommand{\x}{\mathbf{x}}
\newcommand{\y}{\mathbf{y}}

\title{Trainable \& Dynamic Computing: Error Backpropagation Through Physical Media}
\author{Michiel Hermans, Michael Burm, Joni Dambre, Peter Bienstman}

\begin{document}
\maketitle

\textbf{Machine learning algorithms, and more in particular neural networks, arguably experience a revolution in terms of performance. Currently, the best systems we have for speech recognition, computer vision and similar problems are based on neural networks, trained using the half-century old backpropagation algorithm. Despite the fact that neural networks are a form of analog computers, they are still implemented digitally for reasons of convenience and availability. In this paper we demonstrate how we can design physical linear dynamic systems with non-linear feedback as a generic platform for dynamic, neuro-inspired analog computing. We show that a crucial advantage of this setup is that the error backpropagation can be performed physically as well, which greatly speeds up the optimisation process. As we show in this paper, using one experimentally validated and one conceptual example, such systems may be the key to providing a relatively straightforward mechanism for constructing highly scalable, fully dynamic analog computers. }
\\
\\
In a variety of forms, neural networks have seen an exponential rise in attention the last decade. Once deemed an unworkable academic curiosity, neural networks trained with gradient descent are now outperforming other, more classical approaches in a broad number of challenging benchmarks. One dramatic example is their state-of-the-art performance in computer vision \cite{Krizhevsky2012}.  This, including many other examples,  are problems that are considered easy for humans, and hard for conventional computer algorithms. On top of a number of semi-heuristic methods, this result has been obtained using the \emph{backpropagation} algorithm, a method that has been around since the 60's.
\\
One highly interesting set of neural architectures are so-called recurrent neural networks (RNN), neural networks that have temporal feedback loops. This makes them highly suited for problems that have a natural time element, such as speech recognition \cite{Graves2013} and character-based language modelling \cite{Sutskever2011, Hermans2013}. Conventional neural networks may be limited in solving such tasks, as they can only include a finite temporal context (usually a window of data of a fixed duration). RNNs, on the other hand, can -- at least in principle -- have indefinitely long memory. Indeed, it is believed that temporal feedback loops are primary functional components of the human brain. Unfortunately, RNNs are also slow to train compared to more common feedforward neural networks, as they cannot benefit from parallel computing architectures as much. This means that RNNs have only seen a limited impact of the boost of research into neural networks of the last decade.  
\\
Currently, (recurrent) neural networks are implemented on digital devices, mostly for reasons of convenience and availability. At their core, however, they are analog computers, and they come far closer to mimicking the workings of the brain than classic computation algorithms do.  A computer made up of presently available components that would have enough processing power to match that of the human brain is estimated to consume about 100 Megawatts of power (according to estimates of IBM research), a full seven orders of magnitude more than the 20 Watts required by our own brains. If we ever wish to achieve any degree of scalability for devices performing brain-like computation, we will need to embrace physical realisations of analog computers, where information is encoded by real-valued physical variables, and where these data are processed by letting them interact through physical nonlinear dynamic systems. 
\\
One line of research that has been partially successful in accomplishing this goal is that of Reservoir Computing (RC) \cite{Lukosevicius2009}. This paradigm, which combines several previous lines of research \cite{Jaeger2004,Maass2002a}, essentially employs randomly constructed dynamic systems to process time series. The dynamic system is fed with a time-varying signal that needs to be processed. If the dynamic system is sufficiently varied, non-linear and high-dimensional, it will act as an efficient random feature generator, which expands the input signal into a high-dimensional space in which the time series processing problem becomes much easier to solve. All that needs to be optimised is a linear mapping of these features to a desired output signal, which can be performed efficiently with any linear algebra solver. 
\\
The last decade of research into RC has shown a variety of interesting examples of physical implementations of analog processors. It has been shown to work with water ripples \cite{Fernando2003}, mechanical constructs \cite{Caluwaerts2013,Hauser2012}, electro-optical devices \cite{Larger2012,Paquot2012}, fully optical devices \cite{Brunner2013} and nanophotonic circuits \cite{Vandoorne2008, Vandoorne2014}. Despite some successes, the RC concept still faces the problem of being ineffective for tasks that require a great deal of modelling power (e.g., natural language processing, video stream processing, etc.). The main reason being that, due to the random initialisation of the employed systems, any particular feature that needs to be extracted from the input data needs to be present in the system by sheer luck. Even for moderately demanding tasks this requires a disproportionally large dimensionality of the system at hand, up to tens of thousands of state variables in practice \cite{Triefenbach2014}. When the input dimensionality grows, the odds of having the correct features present becomes vanishingly small. By contrast, neural networks trained by the backpropagation algorithm can build the required nonlinear features internally during the training process. 
\\
In recent work we have shown that it is possible to extend backpropagation to models of physical dynamic systems  \cite{Hermans2014}. We showed that it can serve as an efficient automated tool to find highly non-trivial solutions for complex dynamic problems. So far, this work relies on simulation, however. The backpropagation algorithm operates on a computer model of the dynamic system. If we would use it to train physical analog computers, we would face the same scaling issues that we encountered with systems that approach the human brain. 
In this paper, we offer a definition of a dynamic system that encompasses a very broad set of analog computing systems, including all forms of conventional neural networks. Crucially, we show that the backpropagation algorithm can be performed physically on such systems with only minor additional requirements, which strongly reduces the external computational demands of the optimisation process, hence greatly speeding up the training process. 
\section*{Results}
\begin{figure*}[t]
\begin{center}
\includegraphics[width=1\textwidth]{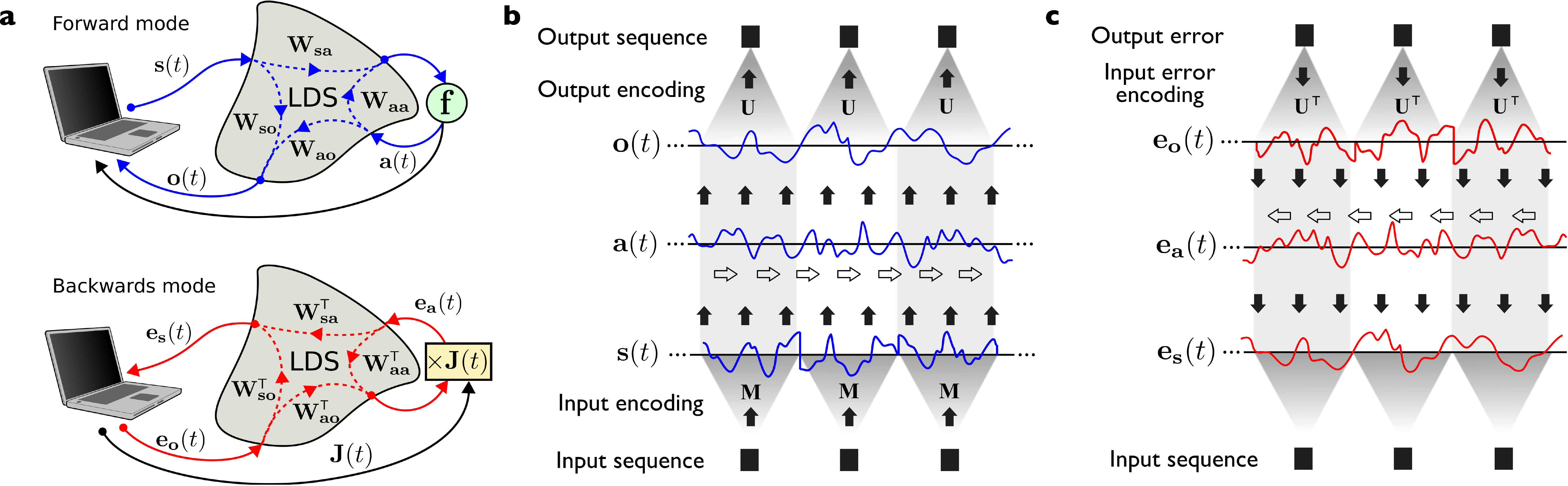}
\caption{\textbf{A:} Illustration of the most general setup of the physical neural network studied in this paper. The top diagram shows how the signals propagate through the system and the nonlinear feedback with blue arrows during the forward pass. The filter operations are depicted with dashed lines running through the linear dynamic system (LDS), which is depicted as a grey blob.  The bottom diagram shows the error backpropagation phase, where the signal runs backwards through all functional dependencies. Here, filter operations run in the opposite direction such that they are represented by the transpose of their impulse response matrices. Note that the computer is not in the loop during the forward or backward pass, but only serves to send out a predefined signal, and to record at the same time. \textbf{B:} Depiction of the masking principle in the forward direction. At the bottom we see three consecutive instances of an input time series. Each of these is converted into a finite time segment through the masking signals $\mathbf{M}(t)$. These segments are next concatenated in time and serve as the input signal $\s(t)$ for the dynamic system (where time runs according to the white arrows), and which in turn generates an output signal $\o(t)$. The output signal $\o(t)$ is divided into finite length pieces, which are decoded into output instances of an output time series using the output masks $\mathbf{U}(t)$.  \textbf{C:} The backpropagation process happens in a completely similar manner as in the forward direction. This time, the transpose of the output masks serve as the encoding masks. Finally, the input error signal $\e_\s(t)$ is also segmented in time before it is used to determine the gradients w.r.t. $\mathbf{M}(t)$}

\label{fig:physical_backprop}
\end{center}
\end{figure*}
We start by introducing an abbreviated notation for a multivariate convolution. If $\x(t)$ is a multivariate signal, and $\W(t)$ a matrix with time-varying elements defined for $t>0$, we define the signal $\y(t)$ as the convolution of $\x(t)$ with $\W(t)$ as follows:
\[\y(t) = \int_0^\infty{dt'\;\W(t')\x(t-t')} = \left[\W*\x\right](t).\]
We will consider systems as follows. We will assume that there are a set of $N$ signal input sources $s_i(t)$ which excite the LDS, and a set of $M$ output receivers which receive an output signal $o_i(t)$. We can write both sets as a single source and receiver vector $\s(t)$ and $\o(t)$, respectively. The LDS will cause the following transformation between the source and receiver:
\begin{equation}
\o(t) = \left[\W_{\s\o}*\s\right](t), \label{Volterra_forward}
\end{equation}
where the impulse response, or first order Volterra kernel $\W_{\s\o}(t)$ characterises the transfer function of the system. Furthermore, we are able to use the LDS reversely, where the receivers act as sources, and the sources as receivers. In this case, $\o(t)$ acts as the input of the system and $\s(t)$ represent the signal receives at the places of the original sources. The LDS is described by:
\begin{equation}
\s(t) = \left[\W_{\s\o}^\textsf{T}*\o\right](t), \label{Volterra_backwards}
\end{equation}
i.e., the operation is reciprocal. One important example of such a system which we will use for the rest of the paper is the propagation of waves through a linear medium. Suppose for instance we have a chamber in which we place a set of different speakers and a set of microphones, the signals the microphones receive would indeed be described by equation \ref{Volterra_forward}, where $\W_{\s\o}(t)$ would be determined by the shape of the room, the absorption of the walls, the air density, etc. If we then replace each speaker by a microphone and vice versa, the signal we received would be described by equation \ref{Volterra_backwards}. Another set of examples are systems described by the linear heat (or diffusion) equation. The source would have a controllable temperature, and the receivers would be thermometers, where the medium causes an operation as described by the above equations. 
\\
Linear systems can only perform linear operations on their input signal. In order for the full system to be able to model non-linear relationships, we add nonlinear feedback. We provide a set of $M_\a$ additional receivers and $N_\a$ sources. The signal that is detected at these new receivers gets sent through a non-linear operator $\f:\mathbb{R}^{M_\a}\rightarrow\mathbb{R}^{N_\a}$ and is fed back into the system via the new sources. We denote the signal after the function as $\a(t)$. The impulse response matrix for the transition from the input sources to the receivers for the nonlinear feedback we denote as $\W_{\s\a}(t)$, and those for the transition from the non-linear feedback sources to the output receivers and non-linear feedback receivers with $\W_{\a\o}(t)$ and $\W_{\a\a}(t)$, respectively. A schematic diagram of the full system is shown in the top of Figure \ref{fig:physical_backprop}. The system is described as follows:
\begin{eqnarray}
\a(t) &= &\f\left(\left[\W_{\s\a}*\s\right](t) + \left[\W_{\a\a}*\a\right](t)\right)\nonumber\\
\o(t) &= & \left[\W_{\s\o}*\s\right](t) + \left[\W_{\a\o}*\a\right](t)
\label{eq:full_system_evolution}
\end{eqnarray}
As we argue in the supplementary material, these equations can easily be identified with neural networks, including all kinds of deep networks, recurrent neural networks, etc. If we wish to use this system as a trainable model for signal processing, we need to be able to define gradients for the parameters we can change. First of all we define a cost functional $C(\o(t))$, a function of the whole history of $\a(t)$ in the interval $t\in\{0\cdots T\}$ that we wish to minimize. Next, we define the partial derivative of $C(\o(t))$ w.r.t. $\o(t)$ as $\e_\o(t)$. The error backpropagation process is then described by the following equations:
\begin{eqnarray}
\e_\a(s) &= &\J^\textsf{T}(s)\left(\left[\W_{\a\o}^{\textsf{T}}*\e_\o\right](s) + \left[\W_{\a\a}^{\textsf{T}}*\e_\a\right](s)\right)\nonumber\\
\e_\s(s) &= & \left[\W_{\s\o}^{\textsf{T}}*\e_\o\right](s) + \left[\W_{\s\a}^{\textsf{T}}*\e_\a\right](s).
\label{eq:full_system_backprop}
\end{eqnarray}
Here, $s  =T-t$, i.e., the equations run backward in time (which in practice means that we play $\e_\o(t)$ backwards as input for the system). From the variables $\e_\a(t)$ and $\e_\s(t)$, gradients w.r.t. all impulse response matrices within the system, and w.r.t. $\s(t)$ can be found.
\\
The above equations describe a continuous-time system. Most time-related machine learning problems work on discrete-time sequences, however. In order to use the system to process discrete time series, we use an input encoding scheme which was first introduced in \cite{Appeltant2011}. Suppose we have an input time series $\x_i$ that we wish to map on an output time series $\y_i$. First of all we define an encoding which transforms the vector $\x_i$, the $i$-th instance of the input data sequence, into a continuous time signal segment $\s_i(t)$: 
\[\s_i(t) = \s_b(t) + \mathbf{M}(t)\x_i\;\;\;\;\;\textrm{for}\;\;\;\;\;t\in\left[0\cdots P\right],\]
where $P$ is the masking period and $\s_b(t)$ is a bias time trace. The matrix $\mathbf{M}(t)$ are the so-called input masks, defined for a finite time interval of duration $P$. The input signal $\s(t)$ is now simply the time-concatenation of the finite time segments $\s_i(t)$\\
The output encoding works in a very similar fashion. If there is a time series $\y_i$ which represents the output associated with the $i$-th instance of the input time series, we can define an output mask $\mathbf{U}(t)$. We divide the time trace of the system output $\o(t)$ into segments $\o_i(t)$ of duration $P$. The $i$-th network output instance is then defined as 
\[\y_i = \y_b + \int_0^P{dt\;\mathbf{U}(t)\o_i(t)},\]
with $\y_b$ a bias vector. The process described here is essentially a form of time multiplexing. The backpropagation phase happens very similarly. Suppose we have a time series with instances $\e_i$, which are the gradient of a cost function w.r.t. the output instances $\y_i$. Completely equivalent to the input masking we can now define the error signal $\e_\o(t)$ as a time concatenation of finite time segments $\e^i_\o(t)$:
\[\e^i_\o(t) = \mathbf{U}^\textsf{T}(t)\e_i.\]
Using this signal as input to the system during backpropagation will provide us with an error $\e_\s(t)$, which in turn provides can be used to determine the gradient for the masking signals $\mathbf{M}(t)$. Note that the signal encoding and decoding happens on an external PC as well, and not physically. These operations can be fully parallelised, however, such that we do not lose the advantages of the processing power of the system. Note that, in practice, we always measure signals using discrete sampling. Similarly, input signals are generated with finite temporal precision. In reality this means that the input and output masking signals also will be defined as a finite set of numbers.
\\
The training process works as follows. First we sample a certain amount of data, run it through the physical system and record the output signals. Next, we compute the cost of the output and construct the signal $\e_\o(t)$, which we then also run through the system physically, and record the signal $\e_\s(t)$ (and $\e_\a(t) if applicable$). Once this is completed, we can compute gradients w.r.t. the relevant parameters (impulse response matrices if applicable, and input and output masks). We then subtract these gradients from the parameters, multiplied with a (small) learning rate. This process is then repeated for many iterations until we reach satisfactory performance or until the experiment stops. 

\subsection*{A real-life acoustic setup}
\begin{figure*}[t]
\begin{center}
\includegraphics[width=0.9\textwidth]{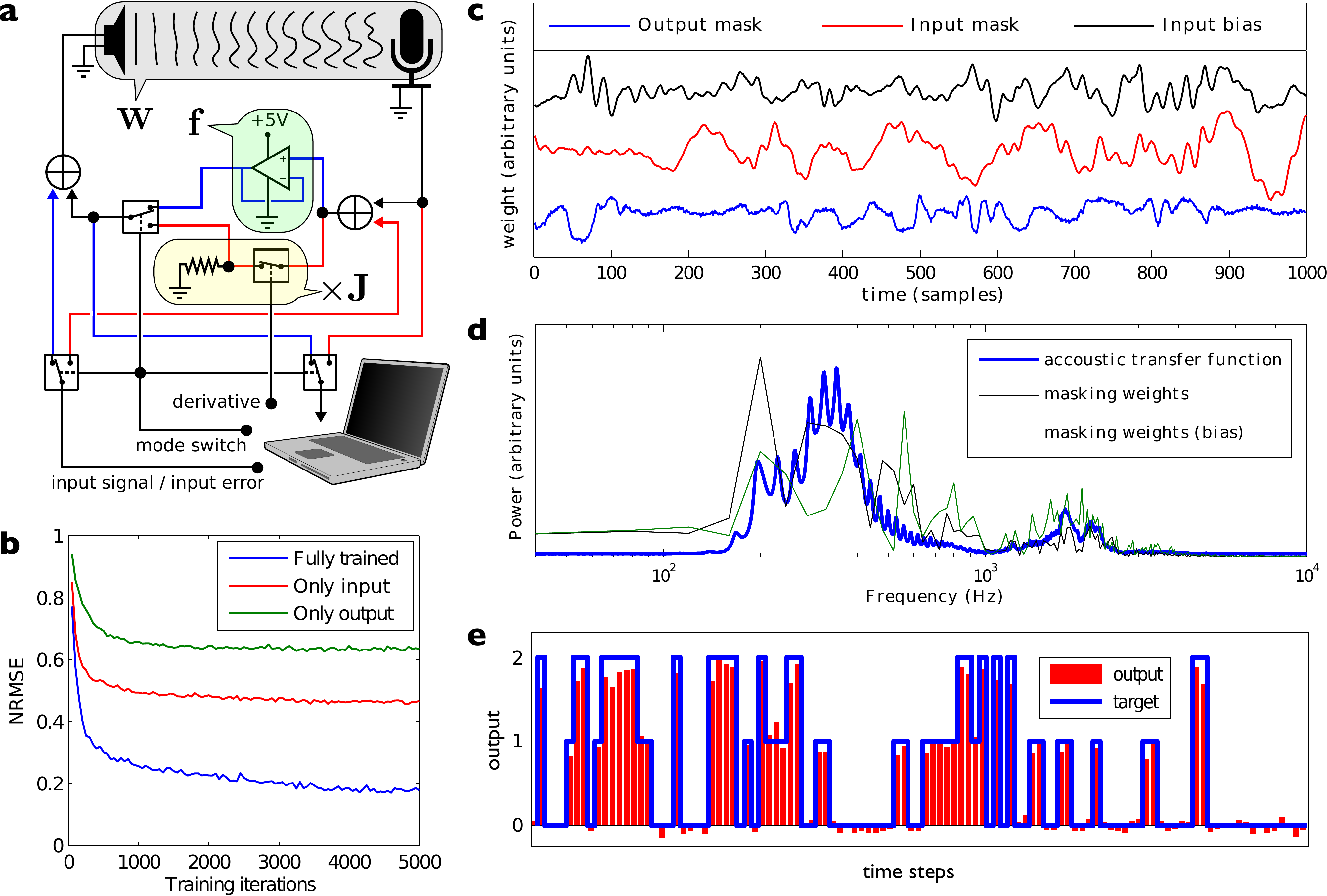}
\caption{Results of the acoustic BPTT experiment. \textbf{a:} Schematic depiction of the electric circuit of the experiment. The grey box at the top represents the speaker, tube and microphone. The blue circuit lines are only in use during the forward propagation, and the red lines are in use during the backpropagation phase. The forward or backward mode can be toggled by a logic signal that comes from the PC, and which controls three analog switches in the circuit. In the forward mode, the nonlinear feedback is implemented by an op-amp voltage follower which cuts off the signal at 0V (green box), implementing a linear rectifier function. In the backprop mode, the multiplication with the Jacobian is implemented by a fast analog switch that either outputs zero, or transmits the signal. \textbf{b:} NRMSE as a function of the number of training iterations for three cases. That where both the output and input masks are trained, and those where only one of each is trained. \textbf{c:} Resulting input and output masks after training. \textbf{d:} Power spectrums of the input masks and the power spectrum of the transmission of the speaker-tube-microphone system. \textbf{e:} Example of the network output vs. the target. }\label{fig:Tube_backprop}
\end{center}
\end{figure*} 

We have tested the principles described above by building a system that uses the propagation of acoustic waves as an LDS. To reduce the complexity of the setup we work with only one signal source (a small computer speaker) and one receiver (a voice microphone). The sound enters a 6-meter long plastic tube via a paper funnel, and the microphone receives the signal at the other end of the tube. The tube will delay the signal and will introduce reflections and resonance frequencies. The received signal is electronically truncated at 0V (such that only positive voltages can pass), implementing a so-called linear rectifier function, which will function as the system non-linearity $\f$. The linear rectifier function is a currently popular activation function in neural architectures \cite{Glorot2011}. Feedback is implemented by adding this signal to the external input signal. One strong advantage of the linear rectifier function is that its derivative is a binary signal: equal to one when the signal is transmitted, and equal to zero when it is cut off. This means that multiplication with the Jacobian is equivalent to either transmitting the feedback signal unchanged, or setting it to zero, which can be easily implemented using an analog switch. For a more detailed explanation of the relation between the acoustic system and the general case described by Equation \ref{eq:full_system_evolution} we refer to the methods section.
\\
We will use the physical BPTT setup to train input and output masks. Note that the output masks are trained simultaneously with gradient descent, but they do not require the physical backpropagation. Both input encoding and output decoding happen on an external PC. 
\\
We tested the setup on an academic task which combines the need for nonlinearity and memory. The input time series $q_i$ is scalar and consists of a series of i.i.d. integers from the set $\{0,1,2\}$, which are encoded to an acoustic signal as described above. The desired output time series $y_i$ is defined as
\[y_i = q(i - q(i)),\]
i.e., the task consists of retrieving the input with a delay that depends on the current input. The fact that the delay is variable makes that the task is nonlinear. i.e., it cannot be solved by any linear filtering operation on the input. 
\\
For details concerning the experiments we refer to the supplementary material. A schematic depiction of the setup and the main results of the experiments are shown in Figure \ref{fig:Tube_backprop}. In Figure \ref{fig:Tube_backprop}e we show a comparison between the system output and the target, indicating that the system has learned to solve the task successfully. We show the evolution of the normalised root mean square error (NRMSE) during the training process in Figure \ref{fig:Tube_backprop}b. To make sure that the physical backpropagation works as intended we have ran two additional tests where we either only train the output masks (the classical RC setup) or only the input masks, keeping the other random and fixed. As can be seen in Figure \ref{fig:Tube_backprop}b, training only output or input masks in both cases reduces the NRMSE. Note that if the input masks are being trained while the output masks are kept fixed and random, all adaptations to the system parameters are exclusively from the error signal that has been propagated through the system in the form of sound. On top of this, the input mask training needs to find a solution that works with a completely random instantiation of the output mask. The fact that it can achieve this at least to some degree (NRMSE $\approx$ 0.47) demonstrates that physical BPTT works as intended. 
\\
The input and output masks are shown in Figure \ref{fig:Tube_backprop}c. In Figure \ref{fig:Tube_backprop}d we show the power spectra of the input masks (and consequently the power spectra of the signals sent into the system) compared to the power spectrum of the system transmission. Clearly, the acoustic parts of the full system (speaker-tube-microphone) only transmit certain frequency bands (a.o. the resonance frequencies of the tube are visible as a set of peaks.), and the input masks seem to have learned to match this spectrum. Note that at no point in training or testing we ever required a model of the acoustic part of the system. The impulse response matrices of the system do not need to be known for the backpropagation to work. Also important to mention is that no process on the external computer or in the electric circuitry provides the required memory for the system to store and process the time series. Instead, this happens due to the memory inherent to the acoustic system and to the nonlinear feedback, which the training process learns to utilise. This means that at any point in time, information about past inputs exists solely as acoustic waves traveling through the tube.

\subsection*{A conceptual optics implementation}
\begin{figure}[t]
\begin{center}
\includegraphics[width=0.4\textwidth]{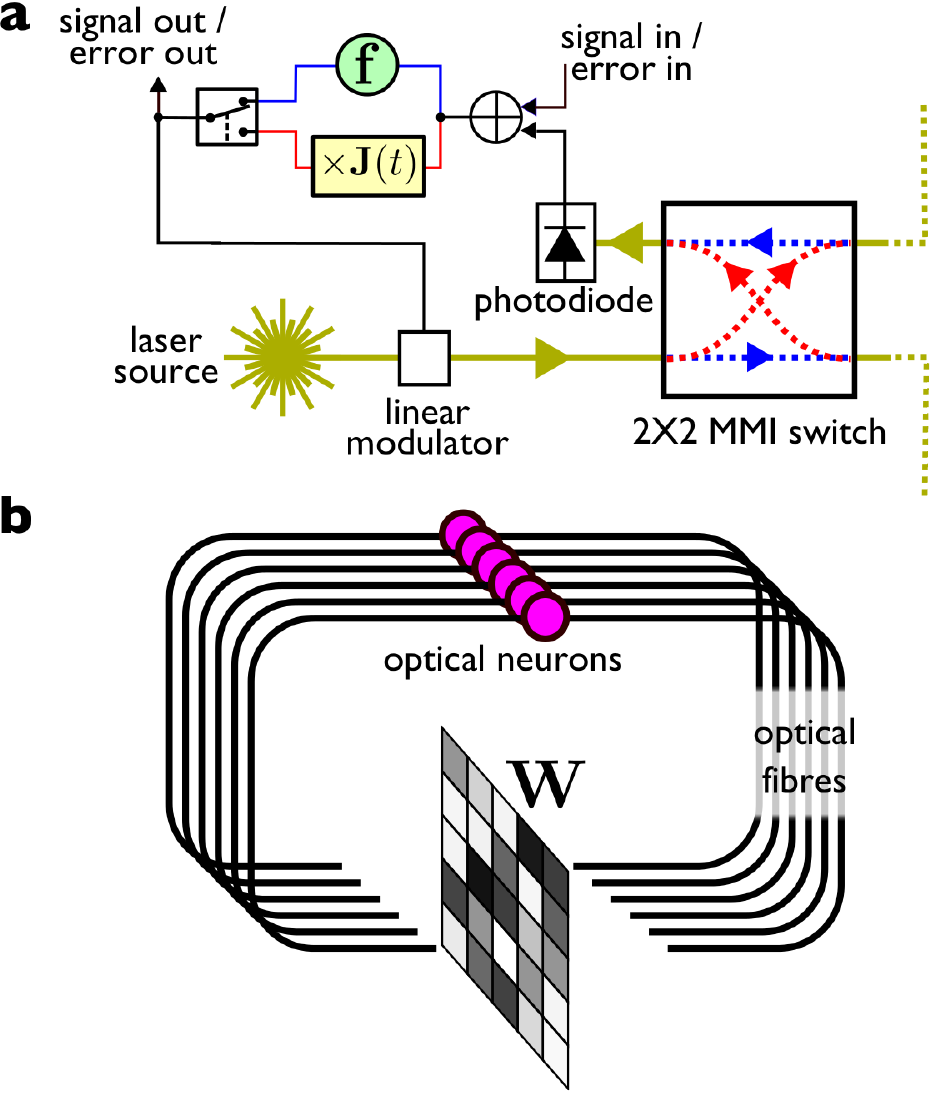}
\caption{ \textbf{a:} Schematic depiction of an optical neuron, where pathways that are exclusive in the forward or backward mode are depicted in blue or red, respectively. It is largely similar to the electronic circuit in the acoustic setup, with the difference that the input now enters the system before the nonlinearity. The 2$\times$2 MMI switch allows light to travel either forwards or backwards through the fibre network (fibres depicted by yellow lines). \textbf{b:} Depiction of a network of optical neurons, Each purple circle represents a neuron. They send their output signals through optical fibres (which also incorporate delay) to an optical matrix-vector multiplier which multiplies with a matrix $\mathbf{W}$ in the forward direction and $\mathbf{W}^{\textsf{T}}$ in the backwards direction. Note that each neuron also has incoming and outgoing connections to external hardware to send input and record output (not depicted). }

\label{fig:optical_neuron}
\end{center}
\end{figure} 

The described acoustic implementation can be extended to a larger and faster system relatively easy. For instance, one could use ultrasound for higher data transmission (combined with a medium with a high speed of sound). One could use piezoelectric transducers that can both emit and receive signals. Sound propagation is in that sense an interesting platform for a physical analog neural network. The most attractive medium, however, but also more technologically challenging, would be light. Light can transport information at a very high speed, and unlike sound waves, it can be easily guided through fiber optics and integrated photonic waveguides. Just like sound waves, light transmission is reciprocal, making it possible to perform error backpropagation physically on the system. Indeed, using light as a medium for neuro-inspired processing has been studied extensively in the past \cite{Caulfield1989,Yu1992}. These examples primarily exploit parallel processing that happens when light travels through a (often holographic) medium. In our case, we would like to exploit not only parallelism, but also the time delays that are inherent to traveling light. 
\\
As a proof of concept we will propose a circuit to perform physical backpropagation optically, partially inspired by the ones described in \cite{Larger2012,Paquot2012}. Delays are physically implemented by means of long optical fibres. For this example we wish not just to train the input masks, but we wish to control the way in which the signals are mixed as well. Concretely, if we have an $N$-dimensional state $\a(t)$, we wish to optically implement a mixing matrix  $\W$ of size $N\times N$, such that the mixing matrix $\W_{\a\a}(t) = \delta(t - D)\W$, where $D$ is the delay introduced by the optical fibres. Setups for computing matrix-vector products optically have been experimentally demonstrated in the past \cite{Goodman1978,Yeh1987}, and here we will assume it is possible to perform them all-optically (see methods). Note that we do not need very high accuracy for neural network applications, as detection of the signal will already be noisy. 
\\
Similar to the acoustic example, we conceive an electro-optical `neuron', which sends and receives optical signals, applies a non-linear function and multiplies with the Jacobian. We encode the state $\a(t)$ as light intensity. Each neuron will have a fixed-power laser source, which can be modulated between a minimal and maximal value. The nonlinear function can be simply implemented by truncating the signal between the minimum and maximum intensity levels of the laser, which conveniently makes full use of the signal range. The Jacobian of such a function is a again a simple binary function, which can be readily implemented electronically. Finally, we can send in the light into the optical circuit either in the forward and backwards direction by using a $2\times 2$ MZI optical switch
\\
The final setup we simulate 20 optical nodes, connected with fibres that incorporate a delay of 109 measurement samples. We add noise to the measured intensity, which gives a signal-to-noise ratio of about 18 dB. For the backpropagation phase, we encounter the problem that the truncation at the minimal and maximal intensity still occurs. If we use a small amplitude for the input error signal, such that the truncation levels are not reached, the signal may be overpowered by the system noise. On the other hand, if we make the magnitude too large, the error signal would be incorrectly truncated. We used a scaling factor that balances these two effects. Some truncation still occurs, but it did not significantly impair the training process. 
\\
We applied the model of this system on a realistic phoneme recognition task, which is part of speech processing and hence a typical example of a problem that can be solved using recurrent neural networks. We used the often-used TIMIT dataset, which is a speech corpus where the phonemes are directly labeled on the speech signal itself. As an error measure, we use the frame error rate, which is the fraction of frames (time steps) in the input signal that have been mislabeled. More details can be found in the methods section. We found that the system is capable of achieving an error rate of 29.6\%. Comparing this to other results in literature we find that we don't perform particularly well, but are in the same ballpark as some other established machine learning techniques (see for instance an overview of results on frame error rate in \cite{Keshet2011}, where the results range from 25\% to 39\%). When we compare to the RC paradigm, similar results are only obtained using extremely large reservoirs, with up to 20,000 nodes \cite{Triefenbach2010}. 
\\
The numerical simulation took roughly 48 hours to finish on a high end processor, and about 95\% of this time was spent on simulating the network, where the rest is mainly spent on computing the parameter gradients. We can roughly estimate how fast the physical setup can be. If we assume that input and output is performed at a speed of one Gigasample per second, which is achievable in commercially available DACs and ADCs, the total time required for gathering all the physically measured data for training the system would take only about 100 seconds, a speedup of three orders of magnitude. Notice that this means that in the current way the training was performed, almost all computation time would be spent on an external processor that computes parameter gradients from the recorded batches of data. Gradient calculations only involve matrix-matrix multiplications, which can again be greatly accelerated using massively parallel processors such as GPUs. Also, the speed at which we can obtain data would make second order methods, such as the Hessian-free approach \cite{Martens2010} highly attractive. This method requires only a small amount of updates, but it does require that gradients are computed on as much data as possible. Finally, the speed of data acquisition can help us tackle noise induced problems. If there is a substantial amount of noise present in the measurements, we can simply repeat the measurement several times and average out the results.

\section*{Discussion}
In this paper we have proposed a framework for using reciprocal physical dynamic systems with nonlinear feedback as analog recurrent neural networks. We demonstrated that the error backpropagation algorithm, which efficiently optimises a recurrent neural network, can be implemented physically on the same system, thus greatly reducing the necessary computations required for the optimisation process, hence speeding up the training process manifold.\\
We have experimentally verified the proposed system using an acoustic setup, where signal processing is performed using sound waves travelling through a plastic tube, and a single nonlinear electronic element provides feedback. Importantly, detailed knowledge of the linear part of the dynamic system is not required in order for the backpropagation to work. No matter how complex the transformation of this part of the system is, the reciprocity makes sure that the error backpropagation happens correctly.\\
To our knowledge, the acoustic example shown in this paper is the first demonstrated non-digital example of a neural architecture where both the processing itself as the error backpropagation happens in a completely dynamic fashion. This research therefore provides a big step towards a completely novel form of processing which relies far less on digital systems, and can directly utilise high-dimensional dynamic systems for processing.\\
The second example we considered is a proof-of-concept  design of an electro-optical system. Unlike sound, light can be manipulated and channeled through fibers. We have demonstrated that the error backpropagation algorithm can be used to optimise the linear impulse response matrices of such system, providing a promising framework for the design of fully trainable electro-optical recurrent neural networks.
\\
In the examples we considered in this paper we mainly optimised input encodings. The scalability of this approach is limited, as the encoding still is performed on an external processor. Future research directions therefore should be focused more on the physical implementation itself. Ideally, we would pass the input signal (light or sound) through a programmable medium, i.e., a medium of which we can control the impulse response matrices in high detail. This would offload the computational complexity of the system to the structure of the physical medium that is used, where the previously mentioned bottleneck no longer would occur. \\ 


\section*{Methods} 
\footnotesize
\subsection*{Acoustic experiments}
For the acoustic setup we used a data acquisition card, which samples the signal at 40 kHz (well above the maximum frequency that is still passed through speaker-tube-microphone system). Input and output masks consist of 1000 samples, which means that time series are processed at 40 time steps per second.\\
We train the input and output masks over the course of 5000 iterations. Initial values for the input and output masks are picked by i.i.d. sampling from a normal distribution with zero mean and variance equal to 0.2 and 0.1, respectively. Each training batch consists of a newly generated time series of 100 instances. This means that each training iteration (forward and backward pass) takes approximately 5 seconds, and the complete training takes about 7 hours. 
\\
As absolute scaling of the error signal does not matter for the backpropagation phase, we always normalised and rescaled the error signal before we used it as input. This was to ensure that the signal always remains well above noise levels. Note that this step causes us to lose the absolute magnitude of the gradients. For parameter updates, we therefore normalised the obtained gradients before using them for parameter updates. The learning rate we chose to be 0.25 at the start of the experiment, and we let it linearly decay to zero to ensure convergence at the end. 
\\
Mathematically, the full system can  be described as follows: Let $s(t)$ be the input signal after the encoding,and $a(t)$ the (scalar) state of the system at time $t$, then
\[a(t) = f\left(\left[W * (s + a)\right](t)\right),\]
where $f(x) = \max(0,x)$, the linear rectifier function, and $W(t)$ is the scalar impulse response of the speaker-tube-microphone system, i.e., it is the signal that would be received by the microphone for a Dirac delta voltage impulse for the speaker. The output of the system is also the state $a(t)$. This means that we can relate the acoustic system to the general case of equation \ref{eq:full_system_evolution} if $\W_{\s\a}(t) = \W_{\a\a}(t) = W(t)$, $\W_{\a\o}(t) = \delta(t)$ (the Dirac delta function), and $\W_{\s\o}(t)=0$.
\subsection*{Conceptual photonic setup}
The system is described by the following equation:
\[\a(t) = f(\W\a(t-D) + \s(t))),\]
where $\W$ is the matrix, which is implemented optically (see below)
The function $f(x)$ truncates the signal between minimal and maximal intensity (which we define as -1 and 1, respectively):
\[f(x) = \begin{cases} -1 &\mbox{if } x \leq -1 \\ 
x & \mbox{if } -1<x<1\\
1 & \mbox{if } x\geq1. \end{cases}\]
\\
For what follows we assume that the coherence length of the used light is very short, such that light intensities add up linearly. In literature, several ways to implement optical matrix-vector multiplication have been discussed. One possible method would be to encode the vector $\a$ as light intensities. We can let the light pass through a spatial array of tuneable intensity modulators and focus the light on the other side to perform a matrix-vector product. This would have the important limitation that all elements of $\W$ would be positive, as light intensities can only be added up. This is a strong drawback, as this would mean that the system can only provide weighted averages of the individual states and cannot use and enhance differences between them. Therefore we envision a different approach where each element of $\a$ is encoded by two light signals with intensities: $\mathbf{k}+ \a$ and $\mathbf{k} - \a$, with $\mathbf{k}$  a vector with all elements equal to one. This ensures that the intensities fall into a positive range between 0 and 2 which can correspond to the minimum and maximum output intensity level of each neuron. Now, these constituents are sent to two separate arrays of intensity modulators $\W_1  = \mathbf{K} + \W/2$ and $\W_2 = \mathbf{K} - \W/2$, where $\mathbf{K}$ has all elements equal to one. As all elements of $\W_1$ and $\W_2$ need to be positive, this means that the range of the elements of $\W$ fall within the range $-2$ to 2. If we combine the light signals after the intensity modulation and hence add up the intensities we get: 
\[(\W_1 + \W_2)\mathbf{k} + (\W_1 - \W_2)\a = \mathbf{Kk} + \W\a.\]
The first term will simply introduce a constant bias value which we can remove electronically after measuring. The second term now contains the matrix-vector product where the elements of $\W$ can be both positive and negative.
\\
For the simulations we chose a piecewise constant input signal with a fixed sample period. We also made sure that all delays in the system are integer multiples of this period. This assures that we can treat the system as a discrete-time system where each time step corresponds to the state of the system during a single sample period. Note that we only did this to speed up the simulations (if the delays are non-integer multiples, the simulation would need higher time precision), and this is not a requirement for the actual physical setup. We again used the masking scheme, where each masking period consisted of 100 sample periods. The delay $D$ we chose at 109 sample periods. We used a network of 20 optical neurons.\\
The TIMIT dataset \cite{Garofolo1993} consists of speech, labeled in time by hand. It has a well-defined training and test set, which makes comparison to results in literature possible. Each time step (frame) needs to be labeled with one out of 39 possible phonemes. The input signal consists of a 39-dimensional time series (coincidentally the same dimensionality of the output signal), which encodes the speech signal using MFCCs. For more details on how the data is encoded, please check, e.g., \cite{Triefenbach2010}.
\\
We trained for 50,000 iterations, where each iteration we randomly sampled 200 sequences of 50 frames from the training set. For the full training process this means that we recorded $50000\times 200\times50$ data points, each containing 100 samples, and we did this twice (for both the forward mode and the backwards mode), leading to $10^{11}$ measurements. At a measuring rate of 1 Gigasample per second this constitutes a total duration of 100 seconds.

We again normalised the gradients. Parameter updates were performed with a learning rate starting at 1, which linearly dropped to zero over the course of the training.

\pagebreak
\newpage
\clearpage
\section*{Supplementary material}
\section{Structure}
This document is the supplementary material for the paper: \emph{Scalable, Trainable \& Dynamic Computing: Error Backpropagation Through Physical Media}. This document consists of three parts. First we motivate the use of the dynamic system we use, secondly we derive a way to define gradients for it, and thirdly we explain how to use time-multiplexing to enhance a low-dimensional system into a high-dimensional one. 
\section{Motivation for the convolutional structure}
We will consider systems that are described by the following set of equations:
\begin{eqnarray}
\a(t) &= &\f\left(\left[\W_{\s\a}*\s\right](t) + \left[\W_{\a\a}*\a\right](t)\right)\nonumber\\
\o(t) &= & \left[\W_{\s\o}*\s\right](t) + \left[\W_{\a\o}*\a\right](t)
\label{eq:full_system_evolution}
\end{eqnarray}
Here, $\s(t)$ is the input signal, $\a(t)$ is the internal state of the system and $\o(t)$ is the output of the system, and we've used the following notation for the convolution operation:
 \[\left[\W*\x\right](t) = \int_0^\infty{dt'\;\W(t')\x(t-t')}.\]
By specific choices of the input signal $\s(t)$ and the impulse matrices $\W_{\mathbf{xy}}(t)$, a wide range of common forms of neural networks  are described by this system. First of all, let's consider the case where $\W_{\a\a}(t)=\mathbf{0}$, $\W_{\s\o}(t)=\mathbf{0}$, and $\W_{\s\a}(t) = \delta(t)\W_\s$, $\W_{\a\o}(t) = \delta(t)\W_\a$. The equation reduces to:
\[\o(t) = \W_\a\f\left(\W_s\s(t)\right),\]
which is nothing more than the description of a neural network with one hidden layer. A deep neural network (a multilayered perceptron) with $L$ layers can be constructed also. We choose $\W_{\s\o}(t) = \mathbf{0}$, $\W_{\s\a}(t) = \delta(t)\W_\s$, $\W_{\a\a}(t) = \delta(t)\W_\a$, and $\W_{\a\o}(t) = \delta(t)\W_\a$, with the following respective definitions:
\[\W_\s = \left[\begin{array}{c}\W_{0}\\
\mathbf{0}\\
\mathbf{0}\\
\vdots \\
  \mathbf{0}  \end{array}\right],\]
  
  \[\W_\a= \left[\begin{array}{cccccc}
  \mathbf{0} &  \mathbf{0}   &\mathbf{0} &\cdots & \mathbf{0} &  \mathbf{0} \\
  \W_1 &   \mathbf{0} &  \mathbf{0} &\cdots & \mathbf{0} &  \mathbf{0} \\
  \mathbf{0} &   \W_2 &  \mathbf{0} &\cdots & \mathbf{0} &  \mathbf{0} \\
    \mathbf{0} &   \mathbf{0} &  \W_3 &\cdots & \mathbf{0} &  \mathbf{0} \\
\vdots          & \vdots           & \vdots          &           &\vdots          &\vdots\\         
    \mathbf{0} &   \mathbf{0} &  \mathbf{0} &\cdots & \W_{L-1} &  \mathbf{0} 

\end{array}\right],\]
\[\W_\a =  \left[\begin{array}{cccccc}
  \mathbf{0} &  \mathbf{0}   &\mathbf{0} &\cdots & \mathbf{0} & \W_L
  \end{array}\right]\]

 If we further assume that $\f$ acts as a scalar function on each of the elements of its argument, we can rewrite the total system as:
 \[\o(t) = \W_L\f(\W_{L-1}\f(\W_{L-2}\f(\W_{L-3}\f(\cdots\W_2\f(\W_1\f(\W_0\s(t))))))),\]
which is exactly the description of a deep neural network. Note that we do not need the transition matrices to be instantaneous (defined by delta functions) per se. If we hold $\s(t)$ fixed, the system may have internal delays, and we would simply need to hold the input for a long enough  time for all internal delays to be resolved. \\
The system can quite simply be transformed into a recurrent neural network. We define the input signal to be piecewise constant with period $\Delta$, and during each period the input is the next instance from a time series $\s_i$. We then simply need to define $\W_{\s\o}(t) = \mathbf{0}$, $\W_{\s\a}(t) = \delta(t)\W_\s$, $\W_{\a\a}(t) = \delta(t-\Delta)\W_\a$, and $\W_{\a\o}(t) = \delta(t)\W_\o$. This system will now act as a recurrent neural network, where each period $\Delta$ the system undergoes a state update.
\[\o(t) = \W_\o\a(t), \]
\[\a(t) = \f(\W_\s\s(t) + \W_\a\a(t-\Delta)).\]

As may be inferred from these three simple examples, far more intricate or unusual neural architectures also fit in the framework described by equation \ref{eq:full_system_evolution}. For instance, if we simply keep the input signal constant, i.e., $\s(t) = \s$ and we assume that under this condition, the system will eventually settle into a steady state, the convolution operations simply become matrix multiplications, and the internal state of the system is described by the following implicit equation:
\[\a = \f(\W_{\s\a}\s + \W_{\a\a}\a),\]
which almost never occurs in the neural networks literature, yet poses no problem on our physical setup, and such a system can be trained in the same manner as any system described by equation \ref{eq:full_system_evolution}.

\section{Derivation of gradients}
\begin{figure*}[t]
\begin{center}
\includegraphics[width=\textwidth]{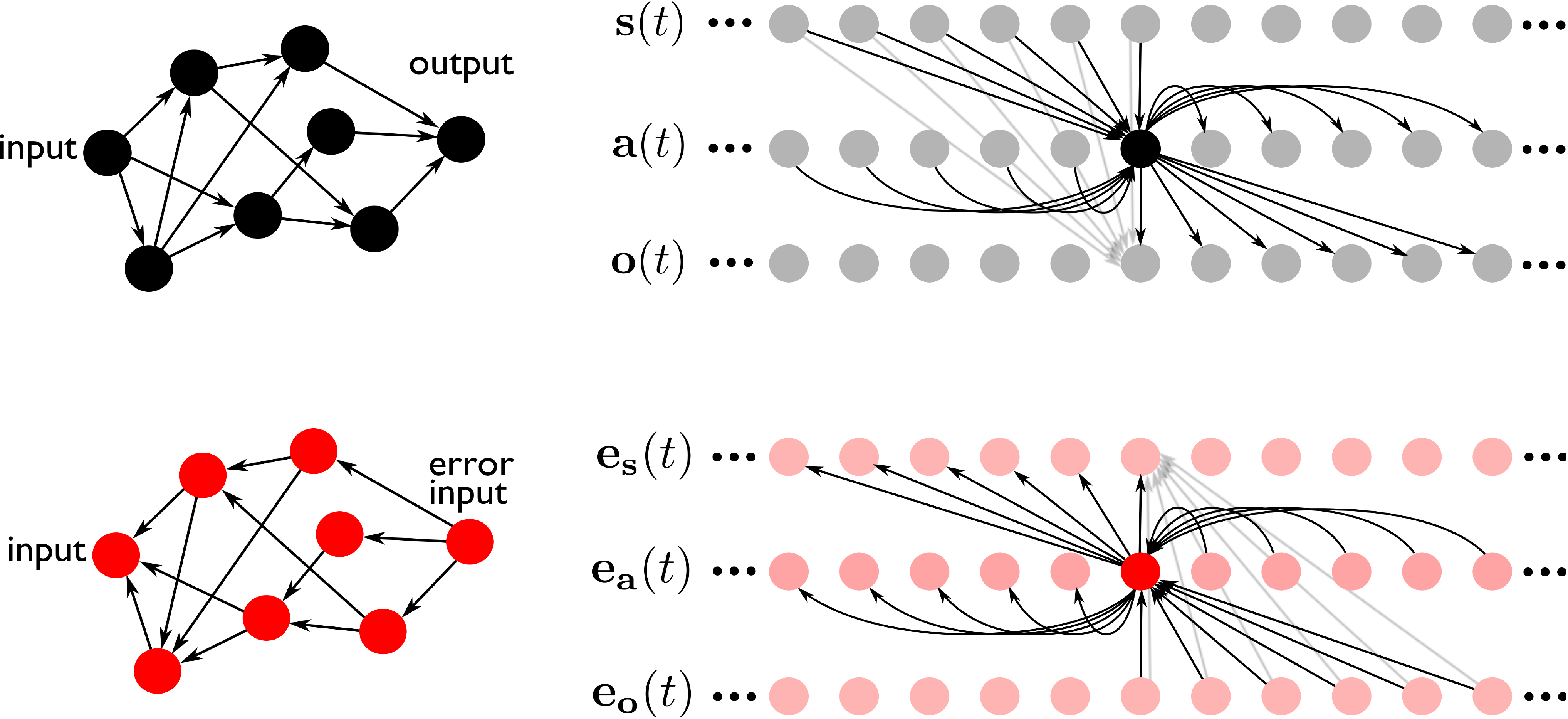}
\caption{Illustration of the graph representation of a neural network architecture. On the left we have depicted the forward and backward graph. Each black circle represents a summation and (optionally) a nonlinear function (except for instance at the output and input nodes). Each arrow represents a linear transformation matrix. During the error backpropagation phase, the direction of  all the edges is reversed, and the error is propagated backwards through the network. Once each node is associated with an error, the transition matrices can be updated. On the right we show the graph associated for a time-discretised approximation of Equation \ref{eq:full_system_evolution}, where we've shown the incoming and outgoing connections of a single node (and in light grey the connections that directly go from the input to the output, represented by $\W_{\s\o}$). Time goes from left to right. Note that, when the directions of the connections are turned around we again end up with a discrete-time convolution.}
\label{fig:neural_graph}
\end{center}
\end{figure*} 
There are several ways to derive the gradients using the error backpropagation mechanism. While it is formally possible to start from equation \ref{eq:full_system_evolution} , it is conceptually easier to use a discrete-time approximation of the convolutions, and consider equation \ref{eq:full_system_evolution} to be a highly specific case of a neural network. First we will give a very brief explanation of the backprop algorithm for an arbitrary neural architecture, and next we consider the special case of the convolutional system used in this paper. For an in-depth proof and explanation of the equations we describe below, we refer to, e.g., \cite{Bishop2006}\\
All neural network architectures can be expressed by a directed graph, where each node performs a nonlinear function, and each edge is a weighted connection. When several edges arrive at a node, their values are added up before they enter the node nonlinearity. The error backpropagation algorithm works by first computing the gradient of the output w.r.t, a certain cost function, (which could be called the output error), and next propagating this error backwards through the directed graph by flipping the direction of each edge. Each edge is still a weighted connection, but during the error backpropagation phase, the error at each node is multiplied with the derivative of the nonlinear function (expressing the current sensitivity of the node). Once the error backpropagation phase is completed, each node in the network will be associated with an error value (the gradient of its activation w.r.t. the cost function). The gradient of each weighted connection (edge) now is simply given by the error at the end times the activation at the front. This principle holds true for any kind of directed graph. 
\\
Instead of considering each node in the graph as a single, scalar function, one can easily expand this to multivariate systems. Now, each node represents a high-dimensional state and optionally a nonlinear function, and each edge represents a linear transformation matrix. Backwards propagation over such an edge is simply a multiplication with the transposed of this linear transformation matrix. Instead of multiplication with the derivative, one multiplies with the jacobian. The gradient w.r.t. the connection matrix is now defined as the outer product of the error vector at the end of the directed edge with the activations at the start.\\
Suppose we have an input vector $\s$. We can then formally define any neural architecture as follows:
\[\x_i = \W^{\s}_{i}\s + \sum_k\W_{ik}\a_k\]
\[\a_i = \f_i\left(\x_i\right)\]
\[\o = \sum_k\W^{\o}_{k}\a_k  + \W^{\o\s}\s,\]
where $\a_k$ represents the state of the $k$-th node of the neural graph, and $\o$ is the output on which we define a cost function. We assume that a subset of the transition matrices are equal to zero, such that no loops occur in the computational graph. Note that the transition matrices in these equation represent the \emph{incoming} connections of each particular node in the graph. We define an error $\e_\o$ as the gradient of the cost function $C$ w.r.t. $\o$:
\[\e_\o = \frac{\partial{C}}{\partial \o}\]
We can now define errors for each node in the graph as follows:
\[\e_i = \J_i^{\textsf{T}}\left(\sum_k\W^\textsf{T}_{ki}\e_k + \sum_k{\W_{k}^{\o}}^\textsf{T}\e_\o\right),\]

with 
\[\J_i = \frac{\partial \f_i(\x_i)}{\partial \x_i}.\]
Here, all the transition matrices in these equations represent the \emph{outgoing} connections of each node in the graph, which represents the fact the error propagates \emph{backwards} through the graph. Finally, gradients w.r.t. each transition matrix in the network can be defined as follows:
\[\frac{dC}{d\W_{ij}} = \e_j\a_i^{\textsf{T}},\;\;
\frac{dC}{d\W^{\s}_{i}} = \e_i\s^{\textsf{T}},\;\;
\frac{dC}{d\W^{\o}_{i}} = \bar{\e}\a_i^{\textsf{T}},\;\;
\frac{dC}{d\W^{\o\s}} = \bar{\e}\s^{\textsf{T}}\]

Note that these equations hold true for any neural architecture. If some the transition matrices are always equal (which is the case for the convolution we will use), one simply needs to add up the gradients that result from all the single outer products described above.\\
When making a discrete-time approximation of the convolutions given in equation \ref{eq:full_system_evolution}, it can be reduced to a directed graph with the properties described above. First of all, we use the following approximation for the convolution:
\[\left[\W*\y\right](t) \approx \delta_t\sum_k{\W(k\delta_t)\y(t - k\delta_t)},\]
which converges to the proper convolution for $\delta_t\rightarrow0$. We have represented the associated computational graph in the bottom of  Figure \ref{fig:neural_graph}, where we've depicted the incoming and outgoing connections of a single node (time instance) of the state vector $\a(t)$. Note that the  connections are shared over time, i.e., the incoming and outgoing connection matrices for different time instance of $\a(t)$ are the same. Also the incoming and outgoing connection matrices that connect the nodes representing $\a(t)$ are symmetric, in the sense that the transition matrix connecting the node representing $a(t-d)$ to $a(t)$ is always equal to the transition matrix connecting node $a(t)$ to $a(t+d)$. \\
We will use the graph representation of the time-discretised approximation of Equation \ref{eq:full_system_evolution}. We start by defining $\e_\o(t)$ as the derivative of the cost function w.r.t. $\o(t)$, We time discretise $\e_\o(t)$, and write down the time-discretised version of Equation \ref{eq:full_system_evolution}:
\begin{eqnarray}
\a(i\delta_t) &= &\f\left(\delta_t\sum_k{\W_{\s\a}(k\delta_t)\s((i - k)\delta_t)} + \delta_t\sum_k{\W_{\a\a}(k\delta_t)\a((i - k)\delta_t)}\right)\nonumber\\
\o(i\delta_t) &= & \delta_t\sum_k{\W_{\s\o}(k\delta_t)\s((i - k)\delta_t)} +  \delta_t\sum_k{\W_{\a\o}(k\delta_t)\a((i - k)}\delta_t)
\label{eq:full_system_evolution_discrete}
\end{eqnarray}
We can now simply apply the previously defined general equations for backpropagation through a neural architecture. By switching the outgoing with the incoming connections we can write down:
\begin{equation}
\e_\a(i\delta_t) = \J^\textsf{T}(i\delta_t)\left(\delta_t\sum_k{\W^\textsf{T}_{\a\o}(k\delta_t)\e_\o((i + k)\delta_t)} + \delta_t\sum_k{\W^\textsf{T}_{\a\a}(k\delta_t)\e_\a((i + k)\delta_t)}\right)
\label{eq:full_system_evolution_discrete_backwards}
\end{equation}
When $\delta_t\rightarrow0$, this reduces again to the continuous-time equation for the backpropagation as provided in the main text. When $\s(t)$ is a trainable representation of another kind of signal, we will also need to propagate the error to the input signal $\s(t)$, which is given by:
\begin{equation}
\e_\s(i\delta_t) =  \delta_t\sum_k{\W_{\s\a}^\textsf{T}(k\delta_t)\e_\a((i + k)\delta_t)} +  \delta_t\sum_k{\W_{\s\o}^\textsf{T}(k\delta_t)\e_\o((i + k)\delta_t)}
\end{equation}
This equation too, reduces to the equation given in the main article when $\delta_t \rightarrow 0$. If errors are computed over a time span $t \in\left\{0\cdots T\right\}$, we can define gradients as 
\[\frac{dC}{d\W_{\s\a}(t)} = \int_0^{T-t}{dt' \e_\a(t'+t)\s^\textsf{T}(t')}\]
\[\frac{dC}{d\W_{\a\a}(t)} = \int_0^{T-t}{dt' \e_\a(t'+t)\a^\textsf{T}(t')}\]
\[\frac{dC}{d\W_{\s\o}(t)} = \int_0^{T-t}{dt' \e_\o(t'+t)\s^\textsf{T}(t')}\]
\[\frac{dC}{d\W_{\a\o}(t)} = \int_0^{T-t}{dt' \e_\o(t'+t)\a^\textsf{T}(t')}.\]
Here the integral stems from the fact that we need to sum op all contributions to the gradients, each from different time steps. This summation becomes an integral in continuous time

\section{Input encoding through time multiplexing}
\begin{figure*}[t]
\begin{center}
\includegraphics[width=\textwidth]{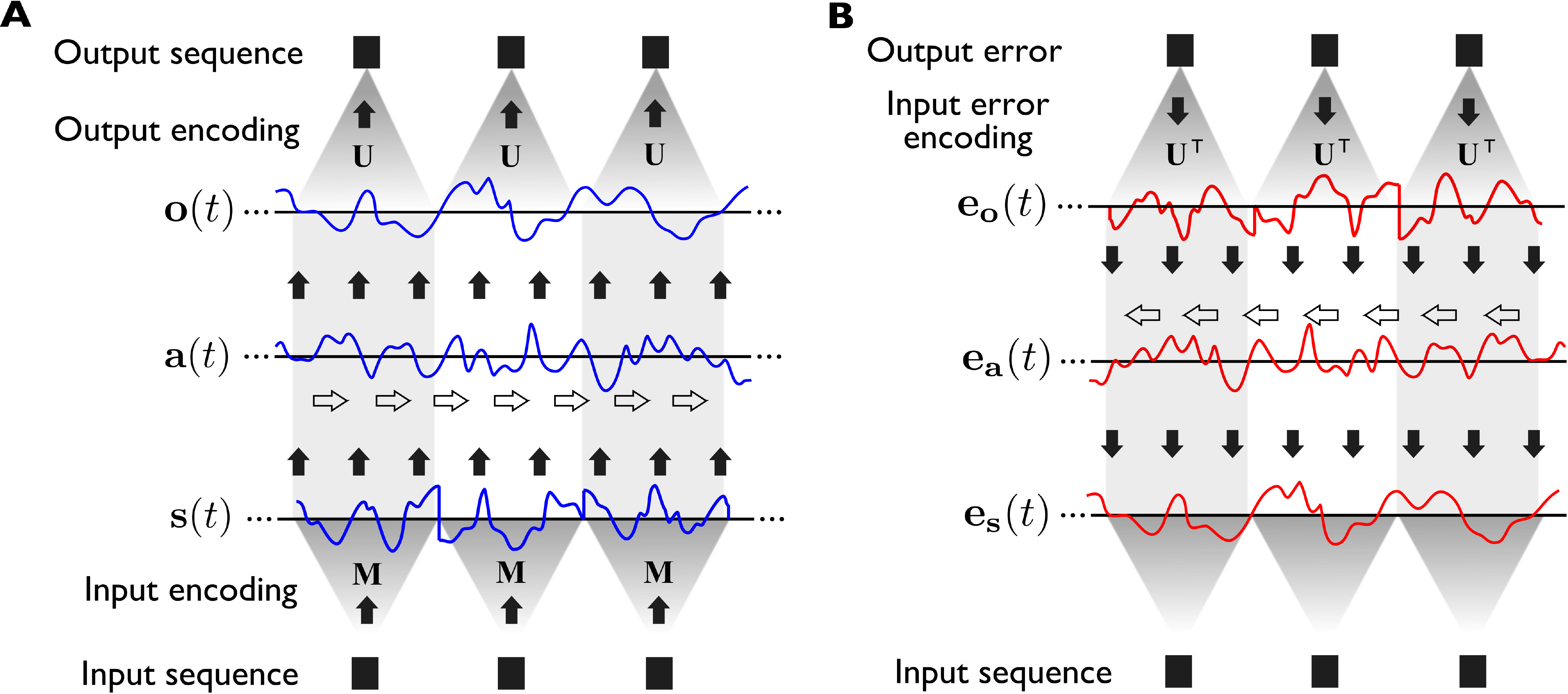}
\caption{Illustration of the masking / time multiplexing principle. \textbf{A:} Depiction of the masking principle in the forward direction. At the bottom we see three consecutive instances of an input time series. Each of these is converted into a finite time segment through the masking signals $\mathbf{M}(t)$. These segments are next concatenated in time and serve as the input signal $\s(t)$ for the dynamic system (where time runs according to the white arrows), and which in turn generates an output signal $\o(t)$. The output signal $\o(t)$ is divided into finite length pieces, which are decoded into output instances of an output time series using the output masks $\mathbf{U}(t)$.  \textbf{B:} The backpropagation process happens in a completely similar manner as in the forward direction. This time, the transpose of the output masks serve as the encoding masks. Finally, the input error signal $\e_\s(t)$ is also segmented in time before it is used to determine the gradients w.r.t. $\mathbf{M}(t)$}
\label{fig:masking}
\end{center}
\end{figure*} 
If we use a physical dynamic system, the input signal we send in is a continuous-time signal. When we wish to process a discrete time series, which is quite typical for many applications,  we need to convert the discrete time signal to a continuous time signal. We will do this using the so-called masking principle, first described in \cite{Appeltant2011}. Here we will use a slightly generalised definition. First of all we define an encoding which transforms the vector $\x_i$, the $i$-th instance of a time series, into a continuous time signal segment $\s_i(t)$: 
\[\s_i(t) = \s_b(t) + \mathbf{M}(t)\x_i\;\;\;\;\;\textrm{for}\;\;\;\;\;t\in\left[0\cdots P\right],\]
where $P$ is the masking period and $\s_b(t)$ is a bias time trace. The matrix $\mathbf{M}(t)$ are the so-called input masks, defined for a finite time interval of duration $P$. The input signal $\s(t)$ is now simply the time-concatenation of the finite time segments $\s_i(t)$\\
The output encoding works in a very similar fashion. If there is a time series $\y_i$ which represents the output associated with the $i$-th instance of the input time series, we can define an output mask $\mathbf{U}(t)$. We divide the time trace of the system output $\o(t)$ into segments $\o_i(t)$ of duration $P$. The $i$-th network output instance is then defined as 
\[\y_i = \y_b + \int_0^P{dt\;\mathbf{U}(t)\o_i(t)},\]
with $\y_b$ a bias vector. The process described here is essentially a form of time multiplexing. The backpropagation phase happens very similarly. Suppose we have a time series with instances $\e_i$, which are the gradient of a cost function w.r.t. the output instances $\y_i$. Completely equivalent to the input masking we can now define the error signal $\e_\o(t)$ as a time concatenation of finite time segments $\e^i_\o(t)$:
\[\e^i_\o(t) = \mathbf{U}^\textsf{T}(t)\e_i.\]
Using this signal as input to the system during backpropagation will provide us with an error $\e_\s(t)$. Exactly equivalent to the output signal, we will now partition $\e_\s(t)$ as a time-concatenation of finite segments $\e_\s^i(t)$. Finally, the gradient w.r.t. the input mask $\mathbf{M}(t)$ is given by:
\[\frac{dC}{d\mathbf{M}(t)} = \sum_i \e_\s^i(t)\x_i^\textsf{T}.\]
For clarity we have illustrated these principles in Figure \ref{fig:masking}. Note that in reality, the signals will be sampled using measurement equipment, and we will define $\mathbf{M}(t)$ and $\mathbf{U}(t)$ as discrete-time signals.

\end{document}